\documentclass[sigconf]{acmart}

\AtBeginDocument{%
  }

\copyrightyear{2024} 
\acmYear{2024} 
\setcopyright{rightsretained} 
\acmConference[ARES 2024]{The 19th International Conference on Availability, Reliability and Security}{July 30-August 2, 2024}{Vienna, Austria}
\acmBooktitle{The 19th International Conference on Availability, Reliability and Security (ARES 2024), July 30-August 2, 2024, Vienna, Austria}\acmDOI{10.1145/3664476.3669926}
\acmISBN{979-8-4007-1718-5/24/07}

\usepackage{dirtytalk}
\usepackage{subcaption}
\usepackage{stfloats}
\usepackage{afterpage}

\begin{document}

\title[Just Rewrite It Again]{Just Rewrite It Again: \\ A Post-Processing Method for Enhanced Semantic Similarity and Privacy Preservation of Differentially Private Rewritten Text}

\author{Stephen Meisenbacher}
\affiliation{%
  \institution{Technical University of Munich \\ School of Computation, Information and Technology \\ Department of Computer Science}
  \city{Garching}
  \country{Germany}}
\email{stephen.meisenbacher@tum.de}

\author{Florian Matthes}
\affiliation{%
  \institution{Technical University of Munich \\ School of Computation, Information and Technology \\ Department of Computer Science}
  \city{Garching}
  \country{Germany}}
\email{matthes@tum.de}

\renewcommand{\shortauthors}{Meisenbacher and Matthes}

\begin{abstract}
The study of Differential Privacy (DP) in Natural Language Processing often views the task of text privatization as a \textit{rewriting} task, in which sensitive input texts are rewritten to hide explicit or implicit private information. In order to evaluate the privacy-preserving capabilities of a DP text rewriting mechanism, \textit{empirical privacy} tests are frequently employed. In these tests, an adversary is modeled, who aims to infer sensitive information (e.g., gender) about the author behind a (privatized) text. Looking to improve the empirical protections provided by DP rewriting methods, we propose a simple post-processing method based on the goal of aligning rewritten texts with their original counterparts, where DP rewritten texts are rewritten \textit{again}. Our results show that such an approach not only produces outputs that are more semantically reminiscent of the original inputs, but also texts which score on average better in empirical privacy evaluations. Therefore, our approach raises the bar for DP rewriting methods in their empirical privacy evaluations, providing an extra layer of protection against malicious adversaries.
\end{abstract}




\maketitle

\section{Introduction}
The proliferation of Large Language Models (LLMs) in recent years has given rise to discussions of data privacy in Natural Language Processing (NLP), particularly as the need for high-quality user-generated text becomes increasingly important to fuel the training of such models \cite{10.1145/3531146.3534642}. While LLMs have demonstrated very impressive capabilities across the spectrum of NLP tasks, the privacy risks inherent in the requirement for massive amounts of training data have motivated the study of privacy-preserving NLP \cite{9152761, klymenko-etal-2022-differential}. This need is made even more salient when considering the sheer amounts of data being passed to hosted LLMs as prompts \cite{edemacu2024privacy}. 

In response to these concerns, a stream of research within the NLP community studies the integration of Differential Privacy (DP) \cite{dwork2006differential} into NLP workflows. As a mathematically grounded blueprint for achieving privacy in data processing scenarios, DP offers a promising solution, yet its direct incorporation into the textual domain does not come without challenges \cite{feyisetan2021research, klymenko-etal-2022-differential, mattern-etal-2022-limits}. Nevertheless, many innovative solutions have been proposed in recent works \cite{hu-etal-2024-differentially}, among these the idea of \textit{differentially private text rewriting}.

In a DP text rewriting scenario, an input text is transformed under DP guarantees with the help of a \textit{rewriting mechanism}, which ideally outputs a privatized text that is semantically similar, yet obfuscated from the original \cite{mattern-etal-2022-limits}. These mechanisms often operate at the \textit{local} level, where users rewrite their data before releasing it to some aggregator. Different rewriting mechanisms operate at various syntactic levels, such as at the word level \cite{feyisetan_balle_2020, carvalho2023tem}, sentence level \cite{meehan-etal-2022-sentence}, or document level \cite{igamberdiev-habernal-2023-dp}. In any scenario, the level at which a DP rewriting mechanism operates leads to the DP guarantee that is provided: for example, given a DP privatized \textit{sentence} in the local setting, this sentence is indistinguishable from all other sentences within some bound, governed by the DP privacy parameter $\varepsilon$.

An important part of designing DP rewriting mechanisms is the evaluation of its privacy-preserving capabilities. In many recent works, this evaluation takes the form of \textit{empirical privacy} tests, where the DP privatized texts are shown to reduce the ability of an attacker to perform some adversarial inference task, as compared to the non-privatized (unrewritten) baseline. In modeling such adversaries, two basic archetypes have been predominantly used in the literature, namely the \textit{static} and \textit{adaptive} attackers \cite{xu2021density,mattern-etal-2022-limits,utpala-etal-2023-locally}.

As shown by empirical privacy evaluations in the literature, it is often the case that the \textit{adaptive} attacker proves to be a considerably more difficult challenge for DP rewriting mechanisms \cite{utpala-etal-2023-locally}, as this attacker is able to mimic the rewriting process and thereby train a more accurate adversarial model. At the same time, achieving better results in empirical privacy evaluations often necessitates lower (stricter) $\varepsilon$ values, which in turn can lead to loss of utility, i.e., semantic similarity to the original texts. This highlights a major challenge of DP text rewriting, that is, finding the balance between privacy and utility \cite{igamberdiev-habernal-2023-dp,meisenbacher2024comparative}.

In this work, we aim to address the challenge presented by the strong capabilities of adversaries modeled in the literature, particularly with the adaptive attacker. To accomplish this goal, we aim to leverage an important property of DP, the \textit{post-processing} principle, to propose a method in which users can enhance the privacy preservation of their DP rewritten texts. As such, we pose the following research question:
\begin{quote}
    \textit{How can users in the local differentially private text rewriting scenario leverage language models to enhance both the empirical privacy and semantic similarity of their rewritten texts?}
\end{quote}

To answer this question, we design, formulate, and evaluate a post-processing method that essentially rewrites the DP rewritten text \textit{again}, with the goal of increasing its privacy while also aligning its semantics to the original text. In evaluating our proposed method, we observe that this extra post-processing step provides clear and significant privacy gains, while also often resulting in higher semantic similarity to the original text counterparts.

We make three contributions to the field of DP text rewriting:
\begin{enumerate}
    \itemsep 0em
    \item To the best of the authors' knowledge, this is the first work to leverage the post-processing property of DP to improve the privacy and quality of DP rewritten texts.
    \item We present a mechanism-agnostic method which demonstrates strong capabilities to enhance the privacy preservation and semantic similarity of DP rewritten texts.
    \item We add to the body of knowledge on DP text rewriting evaluation by highlighting the usefulness of a post-processing step in enhancing the abilities of existing DP mechanisms.
\end{enumerate}

\section{Foundations}
In order to ground our work in previous literature, we now walk through key foundational concepts, which become important in motivating our proposed method.

\subsection{Differentially Private Text Rewriting}
The goal of differentially private text rewriting is to rewrite a sensitive input text in a manner that satisfies Differential Privacy (DP) \cite{dwork2006differential}. Specifically, a mechanism $\mathcal{M}$ with privacy parameter $\varepsilon$ satisfies DP if for any two \textit{adjacent} inputs $x$ and $y$, and $\varepsilon > 0$:
\begin{equation}
    \label{eq:dp}
    \frac{Pr[\mathcal{M}(x) = z]}{Pr[\mathcal{M}(y) = z]} \le e^{\varepsilon}
\end{equation}
In essence, the inequality in Equation \ref{eq:dp} necessitates a certain level of \textit{indistinguishability} between the outputs of two neighboring inputs. The notion of \textit{adjacent} or \textit{neighboring} is dataset-specific, but must be defined in order to satisfy the original notion of DP as defined in Equation \ref{eq:dp}. This level is governed by the $\varepsilon$ parameter: a higher $\varepsilon$ requires less indistinguishability, and vice versa.

The primary challenge with DP text rewriting comes with the design of the underlying \textit{mechanism} that performs the rewriting \cite{klymenko-etal-2022-differential, hu-etal-2024-differentially}. In light of the considerations required by DP, important design decisions include the definition of what \textit{any two text inputs} means, often referred to as \textit{adjacency}. In the literature, adjacency is often either defined on the token-/word-level or the sentence-/document-level. Both approaches have advantages and drawbacks: word-level approaches suffer from lack of contextualization \cite{mattern-etal-2022-limits}, while offering the ability for tighter privacy guarantees. Mechanisms operating on entire documents can better preserve syntactic and semantic coherence, but document representations in large dimensions often necessitate high levels of noise to achieve the DP guarantee, due to their large \textit{sensitivity} \cite{igamberdiev-habernal-2023-dp}.

In this work, we focus on both approaches, namely ones that provide either word- or document-level DP guarantees. We are thereby able to evaluate the effectiveness of our proposed evaluation on both streams of DP text rewriting research. The exact mechanisms we utilize will be introduced in Section \ref{sec:setup}, but firstly, two important notions associated with DP text rewriting are introduced.

\paragraph{Local Differential Privacy}
DP rewriting mechanisms often leverage the \textit{local} notion of Differential Privacy. As opposed to \textit{global} or \textit{central} DP, Local DP (LDP) places the utilization of DP mechanisms at the user level \cite{10.1145/3183713.3197390}. In other words, the transformation (rewriting) of text data is performed by the user locally before releasing the output to some central aggregator. The definition of $\varepsilon$-LDP is as follows, for some mechanism $\pi$, \textit{any} inputs $x$ and $x'$, and $\varepsilon > 0$,
\begin{equation}
    \label{eq:ldp}
    \frac{Pr[\pi(x) = z]}{Pr[\pi(x') = z]} \le e^{\varepsilon}
\end{equation}

As one may observe, the important difference that comes with LDP is the requirement for \textit{any two} inputs from the user to satisfy Equation \ref{eq:ldp}. Although LDP allows for privacy protection to be ensured already at the user level, the major drawback comes with this strict adjacency requirement, thereby often leading to higher amounts of additive noise needed, including in the text rewriting scenario \cite{igamberdiev-habernal-2023-dp}. In this work, we seek to improve this trade-off, whereby our proposed method offers better utility than the raw rewritten outputs of DP text rewriting mechanisms.

\paragraph{Post-processing}
An important property of Differential Privacy that is leveraged in this work is the notion of \textit{post-processing}. In particular, mechanisms that satisfy DP (i.e., satisfy Equation \ref{eq:dp}) are robust against post-processing, defined as any arbitrary operation or computation performed on top of the output of a DP mechanism is safe. Specifically, given a DP mechanism $\mathcal{M}$, some deterministic or randomized function $g$, and input $x$, $g(\mathcal{M}(x))$ upholds the guarantee provided by $\mathcal{M}$ \cite{dwork2006differential}.

This important property states that the output of any function $g$ is still resistant against adversaries, despite these adversaries possessing some auxiliary information. From an information theory perspective, this is due to the fact that regardless of what \textit{auxiliary} knowledge may be possessed by an adversary, lacking knowledge of the private database (here, texts) makes it impossible to compute the function of the outputs of $\mathcal{M}$ \cite{dwork2006differential}. As stated by \citet{near_abuah_2021}, DP mechanisms often leverage this useful DP property to improve the accuracy of DP outputs. As such, we aim to perform post-processing on DP rewritten texts with the goal of enhancing the utility and usability of these texts. 

\begin{figure}
    \centering
    \includegraphics[scale=0.57]{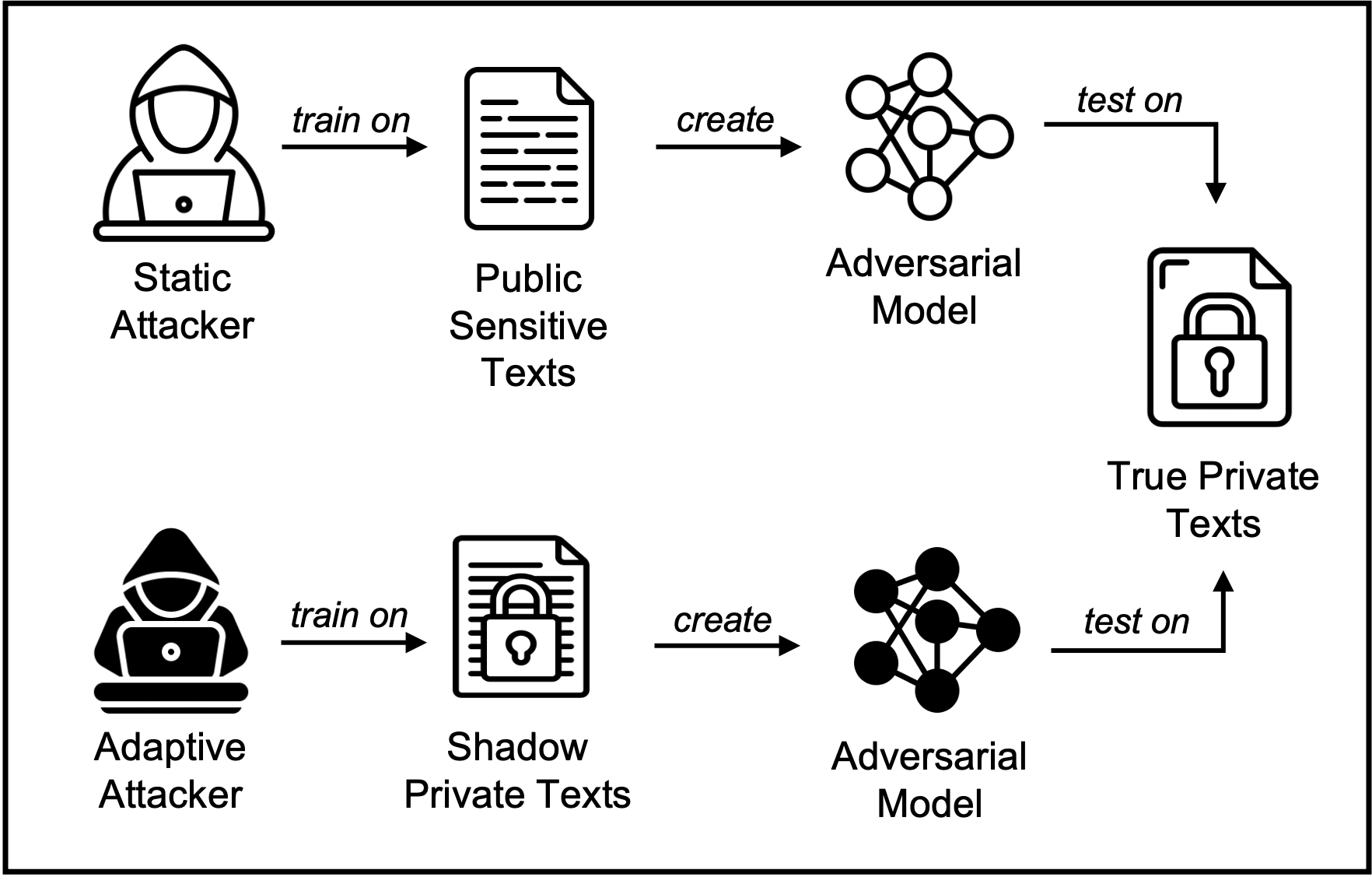}
    \caption{The \textit{Static} and \textit{Adaptive} Attackers. The adaptive attacker, with knowledge of the rewriting mechanism, generates ``shadow'' texts by rewriting publicly available texts from the same domain as the target ``true'' private texts.}
    \label{fig:sa}
\end{figure}

\subsection{Empirical Privacy Evaluation}
\label{sec:ep_explain}
The evaluation of DP text rewriting mechanisms offer takes the form of two-part experiments, resting the \textit{utility} and \textit{privacy} preservation of a proposed mechanism. In particular, a good rewriting mechanism not only protects the privacy of a text, but only rewrites the original text in a way that preserves its downstream utility, i.e., its ability to be useful for training models in some defined task \cite{mattern-etal-2022-limits}.

With this, the question becomes how to evaluate the \say{privacy preservation} of a rewriting mechanism. In many recent works, such an evaluation takes the form of \textit{empirical privacy} experiments, where the demonstration of a mechanism's privacy-preserving capabilities is performed empirically \cite{hu-etal-2024-differentially, mattern-etal-2022-limits, utpala-etal-2023-locally}. At a high level, this is typically done by showing that the output texts (i.e., post-rewriting) reduce the adversarial advantage of an attacker seeking to misuse the text for some nefarious purpose. Prominent examples, which we employ later in this work, include inferring the gender or authorship of the writer behind a given text. As such, a good DP rewriting mechanism should produce texts that reduce an attacker's ability to perform such inferences accurately.

To perform these empirical privacy experiments, two types of attackers have been modeled by the recent literature: the \textit{static} and \textit{adaptive} attackers \cite{mattern-etal-2022-limits}. Both types of attackers work towards the ultimate goal of accurately performing inferences (for a sensitive attribute such as gender) given some corpus of DP-rewritten texts:

\begin{itemize}
    \itemsep 0em
    \item \textbf{Static attacker}: the static attacker has access to the DP rewritten texts, but has no knowledge of the mechanism used to perform the rewriting. The attacker does, however, possess knowledge of the \textit{domain} of the original data, and furthermore, has access to a public dataset of (unrewritten) texts associated with the target sensitive attribute. Using this public data, the static attacker trains a model to predict the target attribute given an input text, and uses this model to infer the attribute of each text in the DP rewritten corpus.
    \item \textbf{Adaptive attacker}: the adaptive attacker possesses all the knowledge that the static attacker does. In addition, the adaptive attacker is stronger in the sense that the exact mechanism (including $\varepsilon$) is known. Using this knowledge, which includes the ability to run the mechanism, the adaptive attacker uses the public data to create a DP rewritten version of this data. The attacker then trains a model on the \textit{rewritten} data to predict the target attribute, i.e., to predict the sensitive attribute of the original DP rewritten texts.
\end{itemize}

Both the static and adaptive attacker setups are illustrated in Figure \ref{fig:sa}. As shown unanimously by recent works \cite{hu-etal-2024-differentially,mattern-etal-2022-limits,utpala-etal-2023-locally}, the adaptive attacker proves to be the stronger adversary. This can be explained by the fact that the distribution on which the adversarial model is trained more closely matches that of the target texts.

In this work, we use both of these attacker types to underline our empirical privacy evaluations. Specifically, we introduce a post-processing method which aims to decrease the adversarial advantage of both attackers, in the way that our post-processing method provides an extra layer of protection on top of DP rewritten texts. This method is introduced in Section \ref{sec:model}.

\subsection{Text2Text Generation}
Recent advances in language models have demonstrated the impressive abilities of these models to generate highly coherent and plausible texts given an input prompt \cite{zhao2023survey}. Many (large) language models also show very strong performance when \textit{fine-tuned} for some specific downstream task, which necessitates only further training data upon which a base model can be improved.

A prominent paradigm of language model fine-tuning comes in the form of \textit{Text2Text Generation} \cite{ijcai2021p0612}, which describes the case where there is a full-text input and full-text output. Common tasks employing such as setup include Machine Translation, Text Summarization, Text Simplification, and Paraphrasing, among others. Given \textit{parallel} datasets, modern LMs can be fine-tuned efficiently, where the base model is then made an \say{expert} by being given domain- and/or task-specific training data.

In this work, we leverage the Text2Text fine-tuning setup to envision a post-processing step to \say{rewrite} texts \say{again}, that is effectively to improve the empirical privacy gains of DP text rewriting by once again rewriting these texts. The definition and requirements of such a method are now introduced in the following.

\section{A Post-processing Method for DP Rewritten Texts}
\label{sec:model}
\begin{figure*}
    \centering
    \includegraphics[scale=0.57]{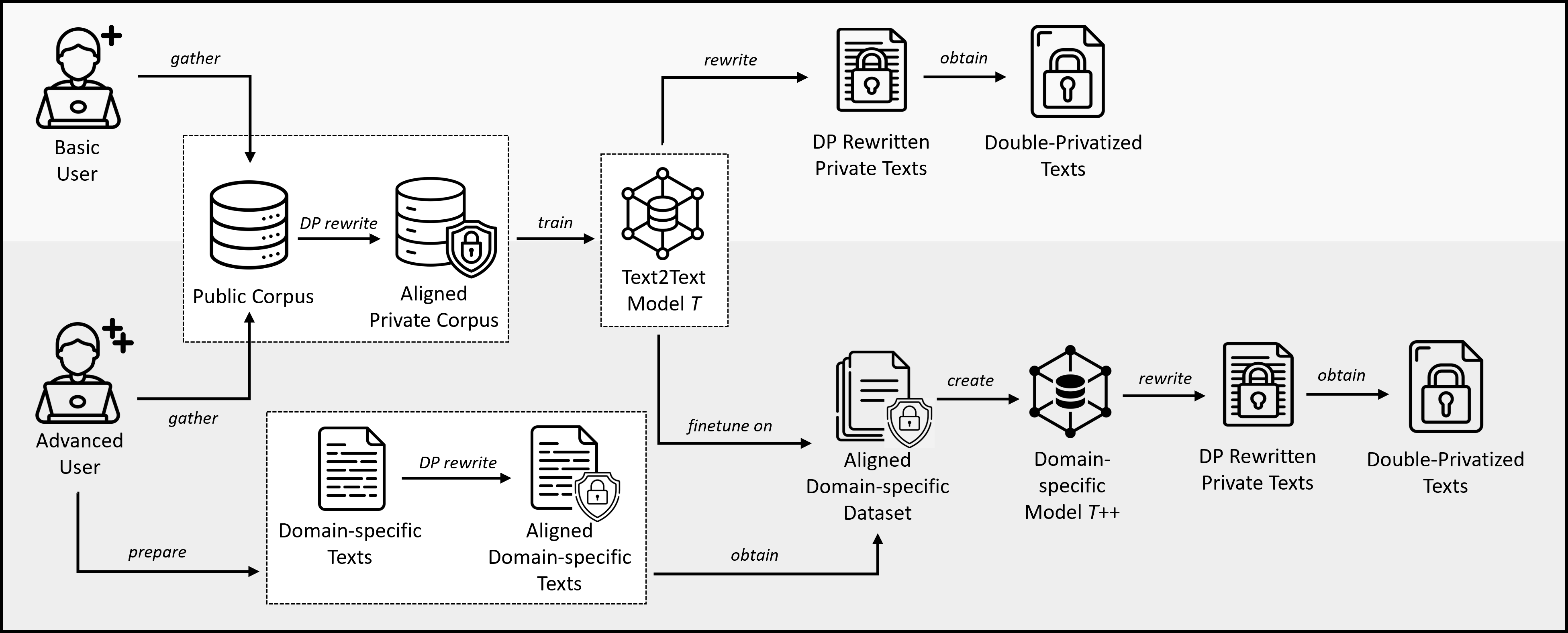}
    \caption{A post-processing pipeline to improve the semantic coherence and privacy preservation of DP rewritten text. Both ``tracks'' involve a user creating a model \textit{T} to rewrite DP rewritten text again. While a basic user utilizes large-scale public text corpora, the more advanced user (bottom track) also leverages domain-specific data to fine-tune \textit{T} further to obtain \textit{T}++.}
    \label{fig:iwaps}
\end{figure*}

In the following, we outline in detail our proposed post-processing method, which can be broken down into two tracks. We present these methods as a way to improve both the semantic similarity and privacy preservation offered by DP rewritten texts. The process flow of these two tracks is illustrated in Figure \ref{fig:iwaps}.

\subsection{Preliminaries}
We assume a user wishing to use some DP rewriting mechanism in the LDP setting. Concretely, a user will rewrite his or her textual dataset before releasing it to some data aggregator or central analyst. To do this, the user leverages a DP mechanism $\mathcal{M}$ with privacy budget $\varepsilon$, where this budget is chosen for the entire text dataset to be rewritten. Additionally, we assume that the user has access to large-scale public text corpora, as well as the ability and resources to fine-tune LLMs on such data in a Text2Text Generation setup.

The motivation of the user is to re-align DP rewritten output text to be semantically closer to the original meaning of the text, thereby boosting the utility and fidelity of the text eventually to be released. As it turns out, an added benefit of performing this extra step also is the enhanced privatization of the output text, providing an added incentive for the user. These benefits will be empirically demonstrated in Section \ref{sec:results} and analyzed in Section \ref{sec:discuss}.

\subsection{Method}
The following outlines the basic steps taken by the user in our proposed post-processing pipeline to rewrite the DP rewritten texts once again. The ensuing narrative then goes into detail on each of these steps. The user takes the following steps:
\begin{enumerate}
    \itemsep 0em
    \item Collect a large number of texts from a public text corpus
    \item Use the rewriting mechanism $\mathcal{M}$ with parameter $\varepsilon$ to rewrite the public corpus into a \textit{aligned private corpus}
    \item Finetune a Text2Text LM to generate text in the \textit{opposite} direction, i.e., generate the public corpus text given the aligned corpus text as input
    \item Use the Text2Text LM to \say{re-rewrite} the DP rewritten texts
    \item Release the doubly rewritten texts
\end{enumerate}

In a more advanced setup, the user takes the same steps (1)-(3), but then continues as follows:
\begin{enumerate}
    \itemsep 0em
    \item[(4)] Generate \textit{aligned private texts} from the same domain as the DP rewritten texts
    \item[(5)] Further finetune the Text2Text LM on this parallel domain-specific data, once again in the \say{reverse} direction
    \item[(6)] Use the further finetuned Text2Text LM to \say{re-rewrite} the DP rewritten texts
    \item[(7)] Release the doubly rewritten texts
\end{enumerate}

\subsubsection{Corpus Creation}
\label{sec:corpus}
The first step of our proposed post-processing pipeline comes with the collection of a large collection of text samples from a publicly available corpus, such as Wikipedia \cite{wikidump} or Common Crawl \cite{JMLR:v21:20-074}. Using these text samples, a user can use the DP rewriting mechanism $\mathcal{M}$ to rewrite the samples in DP rewritten versions, thus creating a parallel, \textit{aligned} corpus.

For the purposes of this work, we utilize a random sample of 100,000 English text samples from C4 (Colossal Cleaned Crawled Corpus), made available by \citet{JMLR:v21:20-074} and cleaned by allen.ai\footnote{\url{https://huggingface.co/datasets/allenai/c4}}.

\subsubsection{LM Fine-Tuning}
Given the prepared aligned corpus, the next task is to fine-tune a Text2Text LM. Much like in the tasks of Machine Translation of Text Simplification, such a fine-tuning task requires a set of parallel \textit{source}-\textit{target} text pairs, which are naturally provided in the DP rewriting process. With this, the DP rewritten texts become the \textit{source} documents, and the original texts represent the \textit{target} documents. This setup is then used to fine-tune an LM, with the effective aim of \say{shifting} the rewritten texts back to the general semantic makeup of the original texts.

\subsubsection{Track Two: Further Fine-Tuning}
As introduced in the above steps, a more advanced user may opt to fine-tune the Text2Text model even further, given the presence of domain-specific texts that more closely mirror the true texts to be released. The motivation behind this is that large-scale public text corpora may contain a vast variety of text subjects, which may not be completely suitable given a set of sensitive texts in one specific domain, that must, in turn, be privatized (rewritten). 

With this domain-specific data, the user would proceed as before with the public corpus, i.e., first generate the parallel, aligned dataset, and then proceed to fine-tune the previously obtained Text2Text model further. The output is thus a model that is more tuned to the target domain, with the ability to recapture the semantics in a particular subject area.

A representative example of domain-specific data, one that we leverage and evaluate later in this work, is that of \textit{user reviews}. If a user is to privatize his or her personal reviews before releasing them, publicly available datasets such as Yelp Reviews \cite{NIPS2015_250cf8b5} or Trustpilot Reviews \cite{10.1145/2736277.2741141} may be useful.

\subsubsection{Data Rewriting and Release}
The final step, given the (domain-specific) fine-tuned LM, is to use the model to rewrite the DP rewritten text outputs. Only these \say{doubly} rewritten text outputs are then shared with third parties, and both the original and DP rewritten texts remain private to the user.

\section{Experimental Setup}
\label{sec:setup}
We describe our experimental setup, which includes the selected DP rewriting mechanisms, design choices for our proposed method, and empirical privacy experiment parameters.

\subsection{Selected DP Mechanisms}
For the evaluation of our proposed method, we choose two DP rewriting mechanisms from the recent literature.

\paragraph{DP-BART}
The \textsc{DP-BART} mechanism \cite{igamberdiev-habernal-2023-dp} was proposed as an LDP mechanism using the BART model architecture \cite{lewis-etal-2020-bart} as its basis. In essence, DP calibrated noise is added to the latent text representation existing between the encoder and decoder components of BART. As such, the rewriting mechanism operates at the \textit{document-level}, where a single output document is generated in a DP manner given an input document. In particular, we utilize the \textbf{DP-BART-CLV} variant as proposed by \citet{igamberdiev-habernal-2023-dp}, which clips the latent vector values before adding noise. 

For $\varepsilon$ values, we choose the first and third quartile values from the range of values evaluated by the original authors. This corresponds to the $\varepsilon$ values of \textit{625} and \textit{1875}.

\paragraph{DP-Prompt}
Leveraging the proxy task of \textit{paraphrasing}, \textsc{DP-Prompt} \cite{utpala-etal-2023-locally} presents a method to generate privatized documents under LDP guarantees by using (large) language models as paraphrasers. Utilizing a temperature sampling mechanism as a form of Exponential Mechanism \cite{4389483}, privatized documents are generated word-by-word by a DP sampling of the next token, thus providing DP guarantees at the \textit{word-level}. For our implementation of \textsc{DP-Prompt}, we use the \textsc{FLAN-T5-large} language model (780M parameters) \cite{JMLR:v25:23-0870}.

As with \textsc{DP-BART}, we use two $\varepsilon$ values as used by the authors of \textsc{DP-Prompt}, namely \textit{137} and \textit{206}. Specifically, following \citet{utpala-etal-2023-locally}, we first measure the logit range of our selected \textsc{Flan-T5-Large} model. This is done by estimating the range via running the model on 100 randomly selected texts from the C4 corpus, and accordingly choosing the clipping range to be $(logit_{min}, logit_{max}) = (-95, 8)$, thus leading to a sensitivity of $103$. Next, we choose the temperature values $T$ of 1.0 and 1.5 (first and third quartile values), which correspond to the two $\varepsilon$ values listed above\footnote{Rounded values from the formula $\varepsilon = \frac{2 \cdot \Delta}{T}$, where $\Delta$ represents the sensitivity.}.

\subsection{Post-Processing Pipeline}
\label{sec:ppp}
\paragraph{Datasets}
As stated in Section \ref{sec:corpus}, we use a random sample of 100,000 text samples from the C4 Corpus to serve as the \textit{public corpus} as envisioned in our post-processing pipeline. Additionally, we use two \textit{domain-specific} datasets, namely the Yelp and Trustpilot reviews, which will be covered in more detail in the following outline of the empirical privacy experimental setup.

\paragraph{Model Fine-Tuning}
In both fine-tuning scenarios, that is the base fine-tuning to create model \textit{T} and the further fine-tuning to obtain \textit{T}++, we once again employ the \textsc{FLAN-T5-large} model checkpoint. Given the input parallel corpus (public-private or domain-specific), the model is trained for one epoch with a learning rate of 5e-5, and otherwise all default HuggingFace Trainer\footnote{\url{https://huggingface.co/docs/transformers/en/main_classes/trainer}} parameters. This process resulted in a total of 12 trained models, namely:

\noindent\textbf{Basic User:}
\begin{enumerate}
    \itemsep 0em
    \item Model \textit{T}, fine-tuned on \textit{aligned private corpus} (\textit{apc}) with \textsc{DP-BART} ($\varepsilon$=625)
    \item Model \textit{T}, fine-tuned on \textit{apc} with \textsc{DP-BART} ($\varepsilon$=1875)
    \item Model \textit{T}, fine-tuned on \textit{apc} with \textsc{DP-Prompt} ($\varepsilon$=137)
    \item Model \textit{T}, fine-tuned on \textit{apc} with \textsc{DP-Prompt} ($\varepsilon$=206)
\end{enumerate}
\textbf{Advanced User:}
\begin{enumerate}
    \itemsep 0em
    \item Model \textit{T}++, further fine-tuned on the Yelp \textit{aligned domain-specific corpus} with \textsc{DP-BART} ($\varepsilon$=625)
    \item Model \textit{T}++, further fine-tuned on the Yelp \textit{aligned domain-specific corpus} with \textsc{DP-BART} ($\varepsilon$=1875)
    \item Model \textit{T}++, further fine-tuned on the Yelp \textit{aligned domain-specific corpus} with \textsc{DP-Prompt} ($\varepsilon$=137)
    \item Model \textit{T}++, further fine-tuned on the Yelp \textit{aligned domain-specific corpus} with \textsc{DP-Prompt} ($\varepsilon$=206)
    \item Model \textit{T}++, further fine-tuned on the Trustpilot \textit{aligned domain-specific corpus} with \textsc{DP-BART} ($\varepsilon$=625)
    \item Model \textit{T}++, further fine-tuned on the Trustpilot \textit{aligned domain-specific corpus} with \textsc{DP-BART} ($\varepsilon$=1875)
    \item Model \textit{T}++, further fine-tuned on the Trustpilot \textit{aligned domain-specific corpus} with \textsc{DP-Prompt} ($\varepsilon$=137)
    \item Model \textit{T}++, further fine-tuned on the Trustpilot \textit{aligned domain-specific corpus} with \textsc{DP-Prompt} ($\varepsilon$=206)
\end{enumerate}

\subsection{Empirical Privacy Experiments}
With the post-processing pipeline, the goal is now to evaluate its effect on the \textit{empirical privacy} protection provided by DP rewriting mechanisms. Additionally, we also measure the effects of the post-processing step on the \textit{utility} of the text, an important counterbalance to be measured in DP text rewriting \cite{mattern-etal-2022-limits, meisenbacher2024comparative}.

\subsubsection{Datasets}
As mentioned in Section \ref{sec:ppp}, we employ two datasets, both of which allow for direct empirical privacy measurement.

\paragraph{Yelp Reviews}
The Yelp Review dataset \cite{NIPS2015_250cf8b5} is a dataset containing user reviews from the popular Yelp platform. Each review contains a star rating from 1-5, leading to the dataset often being used for sentiment analysis. In particular, we use the same data subset as used by \citet{utpala-etal-2023-locally}, which contains 17,295 reviews from 10 frequent users. Thus, we model the empirical privacy task as an \textit{authorship identification} task, wherein an adversary attempts to infer the author given a written text.

\paragraph{Trustpilot Reviews}
The Trustpilot Review dataset \cite{10.1145/2736277.2741141} is a large corpus of reviews from the Trustpilot platform. As with Yelp, each Trustpilot review is scored from 1-5. Additionally, the dataset also lists the gender of each reviewer, thus leading to the \textit{gender identification} adversarial task. As the original dataset is quite expansive, we take a random 10\% of reviews with the gender listed, resulting in an evaluation dataset of 29,490 reviews.

\subsubsection{Modeling Adversaries}
As introduced in Section \ref{sec:ep_explain}, we model two types of adversaries for the empirical privacy evaluation: the \textit{static} and \textit{adaptive} attackers. In order to mimic the data available that these attackers would leverage we perform the following steps:
\begin{enumerate}
    \item For each dataset (Yelp/Trustpilot), we take a 90-10 train-validation split, using a random seed of 42.
    \item The train split is known and used by the attacker. The static attacker only has access to the original version of the texts, while the adaptive attacker has access to the privatized (rewritten) versions.
    \item The validation split represents the \say{true} private text, i.e., the text which the user is releasing, and which the user aims to protect further using post-processing. This text, released in privatized form, is the target of both adversaries.
\end{enumerate}

Given the train split as described above, the attacker in question trains an adversarial model to infer the sensitive attribute, i.e., author or gender. The static attacker trains on the original (clean) texts, while the adaptive attacker trains on the privatized texts (rewritten by mechanism $\mathcal{M}$).

To build these adversarial classifiers, a \textsc{DeBERTa-v3-base} model \cite{he2021debertav3, he2021deberta} is fine-tuned. The model is trained with a 10-class classification head for the authorship identification task, and a 2-class head for gender identification. Training is run for one epoch on a given dataset with a learning rate of 5e-5 and otherwise default training parameters. With the static attacker, a single model is trained and then evaluated on all DP rewritten variants, while for the adaptive attacker, a model is trained for each variant and subsequently evaluated on the corresponding privatized validation set. All model training is performed on a single NVIDIA RTX6000 48GB GPU.

\subsection{Metrics}
Empirical privacy is measured by the ability of a mechanism's rewritten text to \textit{reduce} the adversarial advantage of a given attacker. In this study, this is represented by the reduction in F1 score of the attacker evaluating an adversarial model on the private validation split. This reduction is presented against the baseline of training \textit{and} testing on the clean, non-privatized version of the data, i.e., what an attacker could achieve given full access to the entire target dataset. As such, a lower F1 score represents a higher empirical privacy result. It is important to note that all scores represent an average of three runs, that is, a model is trained three times on different shuffles of the dataset, and the three evaluation results are averaged for the final score.

We also measure the preserved \textit{utility} of the (doubly) rewritten text data, modeled as the \textit{semantic similarity} between the (\textit{original}, \textit{rewritten}) and (\textit{original}, \textit{doubly rewritten}) pairs of text. To measure semantic similarity, we utilize Sentence Transformer models \cite{reimers-2019-sentence-bert}, specifically the \textsc{all-mpnet-base-v2} and \textsc{all-MiniLM-L6-v2} encoder models. The cosine similarity between the encoded pairs of texts is taken and subsequently averaged over an entire rewritten dataset. Since two models are used to account for model-specific differences, the cosine similarity (CS) scores reported represent the average score between the two models.

\section{Empirical Results}
\label{sec:results}

The results of our empirical privacy experiments, namely on the Yelp and Trustpilot datasets, are given in Table \ref{tab:ep}.

As noted in Table \ref{tab:ep}, we present the empirical privacy results in the form of \textit{adversarial advantage}, or specifically the F1 score achieved by an adversary (static or adaptive) in all selected scenarios. The baseline scenario depicts the performance an attacker could achieve given full access to the original, non-privatized data. Then, given a (\textit{mechanism}, $\varepsilon$) pair, we present the empirical privacy results for each rewriting scenario: (1) base output from the DP rewriting mechanism, (2) double-privatized outputs from our \say{basic user}, and (3) double-privatized outputs from our \say{advanced user}.

The results also include the \textit{CS} score, as introduced previously, which captures the degree to which semantic meaning is kept from the original text to the rewritten text. Thus, we pose that a higher \textit{CS} score indicates a closer preservation of original semantic meaning.

\begin{table*}[ht!]
\centering
\begin{subtable}{\textwidth}
\centering
    \resizebox{\textwidth}{!}{
    \begin{tabular}{ccccccccccccc}
    \multicolumn{1}{r|}{Baseline F1} & \multicolumn{12}{c}{\textbf{95.32}} \\ \hline
    \multicolumn{1}{r|}{Mechanism} & \multicolumn{6}{c|}{\textsc{DP-BART}} & \multicolumn{6}{c}{\textsc{DP-Prompt}} \\ \hline
    \multicolumn{1}{r|}{$\varepsilon$} & \multicolumn{3}{c|}{625} & \multicolumn{3}{c|}{1875} & \multicolumn{3}{c|}{137} & \multicolumn{3}{c}{206} \\ \cline{2-13} 
    \multicolumn{1}{l|}{} & F1 (stat.) $\downarrow$ & F1 (adapt.) $\downarrow$ & \multicolumn{1}{l|}{CS $\uparrow$} & F1 (stat.) $\downarrow$ & F1 (adapt.) $\downarrow$ & \multicolumn{1}{l|}{CS $\uparrow$} & F1 (stat.) $\downarrow$ & F1 (adapt.) $\downarrow$ & \multicolumn{1}{l|}{CS $\uparrow$} & F1 (stat.) $\downarrow$ & F1 (adapt.) $\downarrow$ & CS $\uparrow$\\ \hline
    \multicolumn{1}{l|}{Rewritten} & 25.72 & $50.12_{0.8}$ & \multicolumn{1}{l|}{0.31} & \textbf{22.02} & $70.91_{0.6}$ & \multicolumn{1}{l|}{\textbf{0.57}} & 17.92 & $18.84_{0.7}$ & \multicolumn{1}{l|}{0.23} & 19.71 & $19.23_{1.9}$ & 0.44 \\
    \multicolumn{1}{l|}{Basic 2x} & \textbf{18.96} & $\mathbf{26.28_{0.1}}$ & \multicolumn{1}{l|}{0.31} & 27.11 & $\mathbf{39.34_{1.0}}$ & \multicolumn{1}{l|}{0.50} & 10.86 & $\mathbf{17.57_{0.0}}$ & \multicolumn{1}{l|}{0.19} & 13.47 & $17.59_{0.0}$ & 0.41 \\
    \multicolumn{1}{l|}{Advanced 2x} & 25.20 & $33.62_{0.4}$ & \multicolumn{1}{l|}{\textbf{0.42}} & 40.29 & $48.11_{1.2}$ & \multicolumn{1}{l|}{0.53} & \textbf{9.42} & $\mathbf{17.57_{0.0}}$ & \multicolumn{1}{l|}{\textbf{0.38}} & \textbf{12.95} & $\mathbf{17.57_{0.0}}$ & \textbf{0.48} \\
     &  &  &  &  &  &  &  &  &  &  &  & 
    \end{tabular}
    }
\caption{Yelp Empirical Privacy Results.}
\end{subtable}
\begin{subtable}{\textwidth}
\centering
    \resizebox{\textwidth}{!}{
    \begin{tabular}{ccccccccccccc}
    \multicolumn{1}{r|}{Baseline F1} & \multicolumn{12}{c}{\textbf{73.23}} \\ \hline
    \multicolumn{1}{r|}{Mechanism} & \multicolumn{6}{c|}{\textsc{DP-BART}} & \multicolumn{6}{c}{\textsc{DP-Prompt}} \\ \hline
    \multicolumn{1}{r|}{$\varepsilon$} & \multicolumn{3}{c|}{625} & \multicolumn{3}{c|}{1875} & \multicolumn{3}{c|}{137} & \multicolumn{3}{c}{206} \\ \cline{2-13} 
    \multicolumn{1}{l|}{} & F1 (stat.) $\downarrow$ & F1 (adapt.) $\downarrow$ & \multicolumn{1}{l|}{CS $\uparrow$} & F1 (stat.) $\downarrow$ & F1 (adapt.) $\downarrow$ & \multicolumn{1}{l|}{CS $\uparrow$} & F1 (stat.) $\downarrow$ & F1 (adapt.) $\downarrow$ & \multicolumn{1}{l|}{CS $\uparrow$} & F1 (stat.) $\downarrow$ & F1 (adapt.) $\downarrow$ & CS $\uparrow$\\ \hline
    \multicolumn{1}{l|}{Rewritten} & 60.39 & $60.03_{2.2}$ & \multicolumn{1}{l|}{0.36} & \textbf{59.38} & $65.04_{0.1}$ & \multicolumn{1}{l|}{\textbf{0.62}} & 58.53 & $\mathbf{58.08_{0.0}}$ & \multicolumn{1}{l|}{0.22} & 61.99 & $60.37_{1.6}$ & 0.43 \\
    \multicolumn{1}{l|}{Basic 2x} & 55.99 & $\mathbf{58.51_{0.7}}$ & \multicolumn{1}{l|}{0.28} & 59.48 & $61.47_{0.7}$ & \multicolumn{1}{l|}{0.51} & 54.49 & $58.10_{0.0}$ & \multicolumn{1}{l|}{0.16} & 56.63 & $\mathbf{59.91_{1.3}}$ & 0.33 \\
    \multicolumn{1}{l|}{Advanced 2x} & \textbf{55.17} & $59.38_{0.5}$ & \multicolumn{1}{l|}{\textbf{0.42}} & 60.90 & $\mathbf{60.88_{2.0}}$ & \multicolumn{1}{l|}{0.58} & \textbf{49.61} & $58.09_{0.0}$ & \multicolumn{1}{l|}{\textbf{0.36}} & \textbf{54.73} & $60.30_{0.3}$ & \textbf{0.45} \\
     &  &  &  &  &  &  &  &  &  &  &  & 
    \end{tabular}
    }
\caption{Trustpilot Empirical Privacy Results.}
\end{subtable}
\caption{Empirical Privacy Results for Yelp and Trustpilot. \textit{Rewritten} denotes the DP rewritten texts, while \textit{Basic 2x} and \textit{Advanced 2x} denote the result of our proposed post-processing methods (basic and advanced users). \textit{Baseline F1} denotes the adversarial performance on the original, non-privatized texts. For each experiment setting, the best scoring result is \textbf{bolded}. For the adaptive (adapt.) attacker setting, the reported score is an average of three training runs, and the standard deviation is given as a subscript. \textit{CS} denotes the average cosine similarity score between original and rewritten text.}
\vspace{-5pt}
\label{tab:ep}
\end{table*}

\section{Discussion}
\label{sec:discuss}
We now analyze in-depth the empirical results presented in Table \ref{tab:ep}, as well as discuss the merits and limitations of our method.

\subsection{The Benefits of Post-Processing}
An analysis of the empirical results reveals the strengths of our post-processing method, particularly in reducing adversarial performance. Concretely, in the 16 attacker scenarios presented (8x static, 8x adaptive), either our \textit{basic} or \textit{advanced} method leads to the lowest adversarial performance in 13 of the 16 scenarios. In some cases, particularly with \textsc{DP-BART}, the adaptive attacker's performance is nearly 50\% lower as compared to the DP rewritten texts without post-processing. Similarly, our method also proves to be quite useful in reducing the performance of the static attacker.

Looking to both the static and adaptive scores, one can note that only the texts resulting from our methods can truly \textit{neutralize} the adversarial advantage. In other words, these attackers achieved scores equal to or worse than \textit{majority-class guessing}, i.e., simply choosing one class known to be the most frequent. In the Yelp dataset, the most frequent author writes 17.5\% of the reviews, and in Trustpilot, the split is 57.9\%/42.1\% for male/female. As can be seen in Table \ref{tab:ep}, only results from the \textit{Basic} or \textit{Advanced} rewritten texts even lead to F1 scores lower than these majority class percentages\footnote{As majority class guessing will result in zero false negatives but many false positives.}.

A strength of our method comes with the \textit{advanced} setting, which shows in some cases \textit{both} strong protection against adversaries \textit{and} higher semantic similarity to the original texts than only once-rewritten texts. If one assumes cosine similarity (\textit{CS}) to be an indicator of preserved utility, this becomes a quite interesting finding, in contrast to the belief that higher privacy always comes at the trade-off of lower utility. Even where the once-rewritten text scores the highest in terms of \textit{CS}, the empirical privacy gains shown by the advanced double-rewritten text are much more significant than the loss in \textit{CS}, e.g., \textsc{DP-BART} ($\varepsilon=625$) on the Yelp dataset.

\subsection{Basic vs. Advanced}
An interesting point of analysis is the comparison between our two proposed methods, namely the \textit{basic} and \textit{advanced} users. In short, one clear winner between the two methods is not directly discernible from the results, as each showcases particular strengths. 

While the advanced rewritten results achieve the best score most often in terms of empirical privacy scores, the basic rewritten method still outperforms the singly rewritten text more often (6 vs. 3 times), making a case for the \say{simpler} method that does not require extra fine-tuning. Interestingly, the basic rewritten texts achieve the lowest \textit{CS} scores in all tested scenarios.

The promise of the \textit{advanced} user is clear, as discussed above, particularly in its ability to improve privacy and semantic similarity simultaneously. One must keep in mind, however, that this advanced method necessitates the presence of \textit{domain-specific} data to fine-tune the rewriting model. Therefore, the choice between basic and advanced usage of our post-processing method is ultimately contingent upon available resources as well as user preference.

\subsection{A Case for Rewriting Again}
In the above analysis, we pose that the observed benefits of post-processing DP rewritten texts make the case for adopting our proposed pipeline in empirical privacy evaluations. This method, while incurring the cost of extra training on the user side, presents a clear incentive for users in the DP rewriting scenario: the DP guarantee is upheld, while also producing output texts with higher privacy and semantic similarity to the original texts. We also present a method that is \textit{mechanism-agnostic}, meaning that this post-processing method can be run following any DP rewriting process. Moreover, the \textit{basic} scenario, which does not require domain-specific private data, enables the open-sourcing of the proposed post-processing models to allow for private fine-tuning (advanced user).

Beyond this, our analysis of the output texts reveals that we also begin to tackle some overarching challenges of DP text rewriting, namely in the readability and coherence of the output texts. As noted by previous works \cite{igamberdiev-habernal-2023-dp,meisenbacher2024comparative}, DP text rewriting, particularly at higher privacy levels, runs the risk of producing outputs that are incoherent or repetitive. This comes as a side effect of the random noise addition to text representations, an inevitable result in satisfying DP. In performing a post-processing step on top of DP rewritten texts, we aim to alleviate these challenges by producing better semantically aligned texts. This is ensured by the inherent capability of (large) language models to generate such texts. 

To solidify this point, we present selected examples of text outputs in Appendix \ref{sec:appendix}. One can argue that the texts produced by both the \textit{basic} and \textit{advanced} methods appear to be more fluid and coherent, as compared to their singly DP rewritten counterparts.

\subsection{Limitations and Further Considerations}
\label{sec:limits}
The discussion of the merits of our proposed method is not complete without a discussion of its limitations, including those of our evaluation, as well as the potential drawbacks of certain use cases.

Looking to the second rewriting process itself, namely with either model \textit{T} or \textit{T}++, one may observe that the process is dependent on the fine-tuning of a given language model. In this work, we evaluate one particular model, namely \textsc{FLAN-T5-large}, and therefore it remains a point for future work to investigate the effect of model choice (architecture) and model size (i.e., number of parameters) on the method that we describe in this work.

Beyond the choice of rewriting model, the curation of data for the task of post-processing DP rewritten outputs is also seemingly quite important. As shown by our results, the advanced model \textit{T}++, in general, achieves higher \textit{CS} scores than the basic setting, which can be attributed to the fact that the model was further fine-tuned on domain-specific data aligned to the target data. Even before this, the choice of public corpus for the creation of the \textit{aligned public corpus} is also important, as this serves as the basis for both the basic and advanced setups. In this work, we choose the C4 corpus as a reasonable public corpus, but further studies may be well-served to expand this to other text corpora. In addition, the question of \textit{how much} data should be used to fine-tune the models is also not explored in this work, as we simply choose a random 100k sample.

We design our post-processing method to be \textit{mechanism-agnostic}, meaning it can be run on top of any DP rewritten text outputs, as long as the mechanism is known and implementable. Despite this fact, we hypothesize that the nature of a given DP rewriting mechanism can also play a significant role in the effectiveness of the double-privatized outputs. Looking to the results of of evaluation, where we study two distinct mechanisms, one can already see this effect in action. In particular, the empirical privacy results of \textsc{DP-Prompt} tend to be stronger than those of \textsc{DP-BART}, regardless of whether our method is applied or not. This is most plausibly explained by the manner by which each mechanism rewrites, where \textsc{DP-Prompt} models the task as \textit{paraphrasing}, which often results in much shorter output texts that already \say{compress} much of the information of the originals. In contrast, \textsc{DP-BART} often much better mirrors the original text length, offering more space for semantic closeness to the original text -- this is reflected by the generally higher \textit{CS} scores, although this is difficult to equate across different mechanisms with differing effective $\varepsilon$ scales.

A potential concern with arising from the second rewriting of texts comes with the possibility for \textit{loss in factuality}, which comes as a result of both the double rewriting process, as well as from the known ability of language models to \textit{hallucinate} information. The effect of this generation of seemingly plausible, yet potentially factually incorrect, information is outside the scope of our work, but presents an interesting starting point for future works.

A final consideration involves the inevitable fact that our proposed method adds extra overhead to the process of DP text rewriting, a task that can require significant resources even when not using LLMs in the underlying mechanism \cite{meisenbacher2024comparative}. This is especially the case when considering the task placed upon to user to fine-tune a model locally. Although this can be alleviated by open-sourcing the base model \textit{T}, the task of (re)rewriting considerably adds to the overall time requirement of DP text rewriting. Nevertheless, given the results of our empirical evaluations, we pose that the benefits of this extra task can be weighed individually by each user.

\section{Related Work}
The study of Differential Privacy in Natural Language Processing can be traced back to the creation of synthetic Term-Frequency Inverse Document Frequency (TF-IDF) vectors \cite{10.1145/3209978.3210008}, yet the first work on DP in the rewriting scenario was proposed using a generalized form of DP called \textit{metric} DP \cite{fernandes2019generalised}. Since then, several works have been proposed to improve upon the notion of metric DP for NLP, mainly in the study of different metric spaces and distance metrics for word embeddings \cite{feyisetan_balle_2020, xu2020differentially, xu2021density, xu2021utilitarian, yue, Chen2022ACT, carvalho2023tem, bollegala-etal-2023-neighbourhood, meisenbacher20241}.

Works surveying the field of DP in NLP have raised several challenges to the successful integration of DP in NLP tasks \cite{klymenko-etal-2022-differential,mattern-etal-2022-limits}. A more recent survey categorizes the body of work into the characteristics of DP mechanisms, making the major distinction between \textit{gradient perturbation} and \textit{embedding perturbation} methods \cite{hu-etal-2024-differentially}. In these works, the primary challenges of DP in NLP are highlighted, most notably balancing the privacy-utility trade-off, exploring the meaning of the $\varepsilon$ parameter, formalizing what exactly is being protected under a given DP guarantee, and the transparent and reproducible evaluation of rewriting mechanisms \cite{igamberdiev-etal-2022-dp}.

In response to the challenge of producing semantically coherent DP rewritten outputs, recent works have shifted from word-level perturbations to higher syntactic levels, namely the document level. While older works focus on text generation with auto-encoder-type models \cite{Bo2019ERAEDP,Krishna2021ADePTAB}, more recent works have leveraged the generative capabilities of transformer-based language models \cite{mattern-etal-2022-limits,utpala-etal-2023-locally,igamberdiev-habernal-2023-dp}. Still other works focus on the sentence-level \cite{meehan-etal-2022-sentence} or specifically on DP language modeling \cite{shi-etal-2022-selective,wu-etal-2022-adaptive,zhou-etal-2023-textobfuscator}.

Despite the recent wealth of works on DP text rewriting, little to no solutions have been proposed to improve the utility and/or privacy preservation of the privatized texts \textit{post-generation}. In particular, the property of \textit{post-processing} as a potential benefit remains under-researched in NLP applications that integrate DP. It is here where our work is centered, with the goal of providing an intuitive post-processing step for realigning DP rewritten texts while also improving its privacy protection.

\section{Conclusion}
We propose a post-processing method for differentially private rewritten text, which aims to enhance both the empirical privacy and semantic similarity of rewritten text. The evaluation of our methods, both in the basic and advanced user settings, reveals that our proposed pipeline not only offers significant improvements in reducing adversarial advantage, but it is also successful in \say{realigning} the rewritten texts to mirror the original texts more closely. An analysis of the results shows that it is indeed possible to increase both the privacy and utility (CS) of the rewritten texts, thereby presenting a viable method for post-processing DP rewritten texts with the goal of enhancing their strength against capable adversaries.

The limitations of our work as discussed in Section \ref{sec:limits} provide a clear path forward for future research. Concretely, we propose the following studies to build on our work: (1) an investigation of the effect of different (L)LMs in their usage in our post-processing scenario, most notably testing model size and architecture, (2) a study of the extent to which our method works across DP rewriting mechanisms, particularly focusing on potential drawbacks and negative side effects, and (3) continuing research on novel methods for post-processing DP rewritten text with the goal of making these privatized texts both private and useful.

We see our work as an important step in improving the usability of Differential Privacy in NLP, which still largely remains an academic pursuit. By leveraging post-processing and language models \textit{for good}, we hope that future works will follow suit in advancing data privacy in NLP while also ensuring its practical relevance.

\newpage
\begin{acks}
The authors would like to thank Alexandra Klymenko for her valuable contributions to this work.
\end{acks}

\afterpage{
\begin{table*}[hb]
    \footnotesize
    \centering
    \resizebox{0.85\linewidth}{!}{
\begin{tabular}{ll|p{0.7\textwidth}}
 &  & \multicolumn{1}{c}{\textbf{Review}} \\ \hline
\multicolumn{1}{l|}{} & \textbf{Original} & Saget's has been picking up and dropping off my dry-cleaning and this service is pricey - but so worth it as a timesaver for me.  John (the Manager) is terrific and customer service has been great - they bill me monthly and charge it right to my card.  Best of all the cleaning is expertly done and my clothes look great. \\ \hline
\multicolumn{1}{l|}{} & \textbf{Rewritten} & SOMEONE has been doing a lot of work for me, and I have been a customer of his for over a year and a half and this has been a huge loss for me. I have to be honest, this is the first time I've been doing this, and this is what I'm going to be doing for the last time.S has been picking up and dropping off \\ \cline{2-3} 
\multicolumn{1}{l|}{\textsc{DP-BART} ($\varepsilon$ = 625)}  & \textbf{Basic 2x} & S has been picking up and dropping off your used motorcycle parts in North Kingstown NJ. Some folks may have noticed us recommending them via their facebook page but have never seen one of the vehicles they work on and would love to purchase so we contacted him and spoke with the owner. There are some good points about these cars (or bikes in general), but they seemed really nice. When I spoke with S, he suggested that I could order an entire bunch of new parts for each bike I wanted and that my current set-up was perfect. All of that makes me want to buy something from this guy... \\ \cline{2-3} 
\multicolumn{1}{l|}{} & \textbf{Advanced 2x} & S has been picking up and dropping off food, and the service and promptness to put things away - on time. I really wanted this place to do better and ... well, my dad's truck has a faulty gear box that will run on all 3 batteries and this was the perfect solution to take advantage and fix it, but I guess my car is simply not going to stand by for two weeks. \\ \hline
\multicolumn{1}{l|}{} & \textbf{Rewritten} & I have been using this service for years and it has been great for me. The service has been amazing. I have been picking up and dropping off my clothes at this location for years. I love the service and the way it is done. My favorite part is that the service is free of charge.The service is great and I love it.My favorite part of this service is that I have \\ \cline{2-3} 
\multicolumn{1}{l|}{\textsc{DP-BART} ($\varepsilon$ = 1875)} & {\textbf{Basic 2x}} & I have been picking up and dropping off my clothes at the LVAD store since 2008 and it has always been great! They go above and beyond to make sure every customer is taken care of. My favorite thing about their service is that they are always in control of your belongings and if we need additional items they can be reordered, delivered or picked up on time. My only complaint is that sometimes there are times they change the clothing that they will pick and deliver, but it never changes my items. \\ \cline{2-3} 
\multicolumn{1}{l|}{} & \textbf{Advanced 2x} & Awesome service and quality clothes. My favorite part is that they never have a wait time. Love being pick up and dropped off when you need it and the clean of the facility is always appreciated. The service has been great and the staff is super nice. \\ \hline
\multicolumn{1}{l|}{} & \textbf{Rewritten} & local pug-whip \\ \cline{2-3} 
\multicolumn{1}{l|}{\textsc{DP-Prompt} ($\varepsilon$ = 137)} & \textbf{Basic 2x} & he Pugwop Club is happy to present the Brutus Muppets from New York. These cartoon Pugs have been in production since 1985! Their live debut was the New Century Fox TV cartoon series “Bump It Up”. They also enjoyed several notable roles in television shows, including The People’s Love Band, and are now on NBC. In addition to being part of new shows, including “Pugwit,” and “Geek Squad,” they were cast in movies such as Zombies, Monster High, Spy Kids, Wet’n Wild. \\ \cline{2-3} 
\multicolumn{1}{l|}{} & \textbf{Advanced 2x} & I have had so many good meals here that I've always been reluctant to give them a solid review. My boyfriend and I shared a lovely bowl of Thai-Kogi Curry noodle with some curry chicken for lunch. It was an excellent start to our day. We chose rice as the base, which came topped with veggies from their organic veggie patch. This was made into a huge bowl along with plenty of crispy coconut, sweet basil leaves (whole root vegetables) and ginger in curry sauce. Loved it! \\ \hline
\multicolumn{1}{l|}{} & \textbf{Rewritten} & "The Professional Dry Cleaners can help save time!" \\ \cline{2-3} 
\multicolumn{1}{l|}{\textsc{DP-Prompt} ($\varepsilon$ = 206)} & \textbf{Basic 2x} & We are happy to help you. "The Professional Dry Cleaners can help save time! The team is wonderful! They came on the day they needed us and saved me from an unpleasant experience. \\ \cline{2-3} 
\multicolumn{1}{l|}{} & \textbf{Advanced 2x} & Good times, good service, and this is NOT a dry cleaner's. Why wash dirty sheets for me? They come at 10 minutes so there is usually nothing else they need to do, like run errands or pick up the kids in a diaper. When you walk in and walk through the doors you have a feeling that there's really someone there (or the front desk). Once inside, the store has a few rooms with nice linens (pricey if you ask for them) and some old school ambiance and everything you can hope for.
\end{tabular}
    }
\caption{Rewritten Examples from the Yelp Dataset. \textit{Rewritten} denotes the base DP rewritten texts, while \textit{Basic 2x} and \textit{Advanced 2x} denote our basic and advanced post-processing methods, respectively.} 
\label{tab:examples_yelp}
\end{table*}
}
\bibliographystyle{ACM-Reference-Format}
\bibliography{sample-base}

\appendix
\section{Rewriting Examples}
\label{sec:appendix}
Table \ref{tab:examples_yelp} showcases rewritten examples from the Yelp dataset, specifically the eleventh row.

\end{document}